\documentstyle[fullpage,qtree]{article}    
\input{psfig.tex}
\title{An Empirical Comparison of Probability Models \\ for Dependency Grammar
\thanks{This material is based upon work supported under a National
Science Foundation Graduate Fellowship, and has benefited greatly from
discussions with Mike Collins, Dan Melamed, Mitch Marcus and Adwait Ratnaparkhi.  Special
thanks are due to Mike Collins for kindly testing his parser on my dataset.}}
\date{}
\author{Jason M. Eisner \\
CIS Department, University of Pennsylvania \\
200 S. 33rd St., Philadelphia, PA 19104-6389, USA \\
{\tt jeisner@linc.cis.upenn.edu}}

\newcommand{\angles}[1]{{\left\langle #1\right\rangle}}
\newcommand{\chooz}[2]{\mbox{\scriptsize $\left(\begin{array}{@{}c@{}}{#1} \\ {#2} \end{array} \right)$ \normalsize}}   

\newenvironment{prog}{%
       \vspace{-\baselineskip}
       \setcounter{line}{0}
       \begin{small}%
       \begin{tabbing}%
       \hspace{\parindent}\=1234\=123\=123\=123\=123\=123\=123\=123\=123\=123\=\+\kill}%
    {\end{tabbing} \end{small}}
\newcounter{line}
\newcommand{\progline}{\\ \stepcounter{line}{\tiny \theline .}\>}   
\newcommand{\lprogline}[1]{\\ \refstepcounter{line}\label{#1}{\tiny \theline .}\>}  
\newcommand{\nprogline}{\\ \>}              
\newcommand{\comment}[1]{{\scriptsize (* \sl #1 *)}}
\newcommand{\scomment}[1]{\hspace{0.25in}\comment{#1}}   

\begin{document}
\maketitle
\begin{abstract}
\small This technical report is an appendix to Eisner (1996): it gives
superior experimental results that were reported only in the talk
version of that paper, with details of how the results were obtained.
Eisner (1996) trained three probability models on a small set of about
4,000 conjunction-free, dependency-grammar parses derived from the
{\em Wall Street Journal} section of the Penn Treebank, and then
evaluated the models on a held-out test set, using a novel $O(n^3)$
parsing algorithm.  

The present paper describes some details of the experiments and
repeats them with a larger training set of 25,000 sentences.  As
reported at the talk, the more extensive training yields greatly
improved performance, cutting in half the error rate of Eisner (1996).
Nearly half the sentences are parsed with no misattachments;
two-thirds of sentences are parsed with at most one misattachment.

Of the models described in the original paper, the best score is
obtained with the generative ``model C,'' which attaches 87--88\% of
all words to the correct parent.  However, better models are also
explored, in particular, two simple variants on the comprehension
``model B.''  The better of these has an attachment accuracy of 90\%,
and (unlike model C) tags words more accurately than the comparable
trigram tagger.  

If tags are roughly known in advance, search error is all but
eliminated and the new model attains an attachment accuracy of 93\%.
We find that the parser of Collins (1996), when combined with a
highly-trained tagger, also achieves 93\% when trained and tested on
the same sentences.  We briefly discuss the similarities and
differences between Collins's model and ours, pointing out the
strengths of each and noting that these strengths could be combined
for either dependency parsing or phrase-structure parsing.
\end{abstract}

\section{Introduction} 

\cite{eisnercoling} proposed and compared three lexicalist,
probabilistic models of dependency grammar (together with a parsing
algorithm).  The models' relative performance is of interest, as they
reflect different independence assumptions about syntactic structure.

Although \cite{eisnercoling} included an empirical comparison of the models,
results were not complete by press time.  Thus, unfortunately, the
written version of the paper included only the results of a pilot
study based on a small training set.  The results of a larger
experiment were presented at COLING-96 along with the paper.  It is
the purpose of this technical report to describe those significantly
improved results and a few additions.

The organization of the paper is as follows.  For a conceptual
overview of the work, the reader is encouraged first to read
\cite{eisnercoling}.  \S\ref{sec:models}--\S\ref{sec:corpus} detail
the experimental setup: \S\ref{sec:models} specifies the precise
probability models used for the experiments, \S\ref{sec:smooth}
explains how the probabilities were estimated, and \S\ref{sec:corpus}
describes how the training and test data were prepared.
\S\ref{sec:results} gives the experimental results and discusses them.
\S\ref{sec:conclusion} offers some concluding remarks.

\section{Precise formulation of the models}\label{sec:models}

\subsection{Dependency structures}

We begin with a brief review of the terminology.  A {\bf bare-bones
dependency structure} is a sequence or {\bf string} of $n$ words.
Each word has been annotated with a {\bf tag}, which indicates the
word's syntactic or semantic role, as well as a {\bf parent}, which
indicates where the word plays that role.

The parent of word $w$ is usually a pointer or {\bf link} to another
word in the string, which $w$ is said to {\bf modify}.  However, one
word in the string (the ``head'' of the sentence) modifies nothing,
and its parent is said to be the special symbol {\sc eos}, an
end-of-sentence mark that falls just past the end of the string (as
word $n+1$).  In addition, for the dependency structure to be
well-formed, the parents must be assigned in such a way that the links
never cross or form cycles.  Each word is said to be the {\bf head} of
the contiguous substring formed by itself and all its descendants.

\begin{figure}
\centerline{\psfig{figure=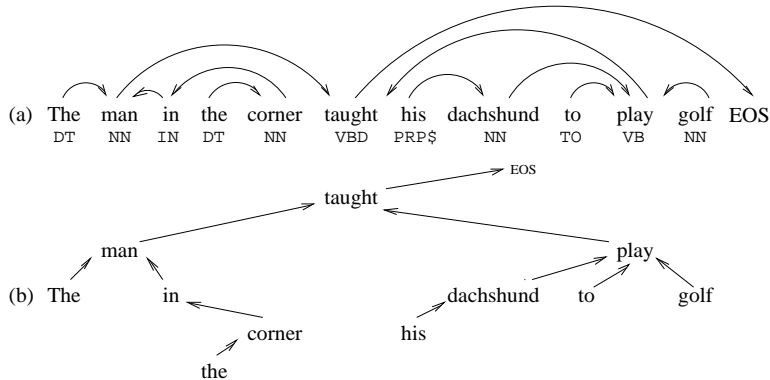,width=4in}}
\caption{{\bf (a)} A bare-bones dependency structure, as described in
the text.  {\bf (b)} Constituent structure and subcategorization may be
highlighted by displaying the same dependencies as a lexical
tree.}\label{sampleparse}
\end{figure}

Figure~\ref{sampleparse} gives an example.  Here the tags are simple
part-of-speech tags as used in \cite{Browncorpus,Treebank}.  It is
possible to use a more articulated tag set, in order to achieve parses
that are more precise and perhaps even more accurate.  For example, in
figure~\ref{sampleparse}, {\em dachshund}'s tag might be extended so
that it includes not only the part of speech, \verb+NN+ (noun), but
also indications that this token of {\em dachshund} heads a definite
NP and serves as a semantic agent.  Aside from questions of smoothing
sparse data, a large tag set can be used without changes to the model.
However, the tag set used in the present experiments is merely a
slightly refined version of the part-of-speech tags of
\cite{Browncorpus,Treebank}.  The refinements are described in
\S\ref{sec:corpus}.

\subsection{Probability models and structural choices}

The present paper compares various {\bf probability models}.  Each
model describes a probability distribution over the space of all
dependency structures for all word strings.  When given a sentence of
words $w_1, w_2, \ldots w_n$, the parser assigns respective tags $t_1,
t_2, \ldots t_n$ and parents $p_1, p_2, \ldots p_n$ so as to maximize
the probability $\Pr(\vec{w},\vec{t},\vec{p})$ of the resulting
dependency structure (words, tags, and all).  That is, the parser
returns the highest-probability structure consistent with the given
word string $\vec{w}$.

Under each model, the probability of a dependency structure $D$ can be
described with reference to a notional process that generates $D$ through
a unique sequence of structural choices.  The probability of occurrence
of $D$ is the probability of making the appropriate sequence of
choices.  Each choice is made randomly, with a probability conditioned
only on certain aspects of the previously made choices.

Certain models do not achieve a 1-to-1 correspondence
between choice sequences and legal dependency structures.  In general,
by assigning probability to a choice sequence, the model assigns a
total probability to the class of all legal structures consistent with
that sequence.  
\begin{itemize}
\item If classes of size $< 1$ exist, the model is called
{\bf leaky} (or deficient): it allocates some probability to
situations that can never legally arise.  
\item If classes of size $> 1$ exist, the model is called {\bf
incomplete}.  It cannot be used to find the probabilities of
individual structures, but only of classes.  The parser therefore
returns not the highest-probability structure, but an arbitrary
representative of the highest-probability class.  (For purposes of
implementation, each parse is scored with the probability of its
choice sequence, so that structures in the same class will suffer
a tie that is broken arbitrarily.)
\item A third possibility is that
the model may be {\bf inconsistent} in that some classes overlap,
i.e., a structure can be generated in more than one way.  In this
paper, however, only consistent models are considered; as stated
above, each $D$ is generated through a {\em unique} sequence of structural
choices.
\end{itemize}
Note that we may eliminate leakiness from a model by renormalizing its
probabilities---that is, by defining $S$ as the total probability allocated to
well-formed structures, and putting
$$\Pr_{Normalized}(D) = \Pr(D \mid D \mbox{ is well-formed}) = \left\{
\begin{array}{cl}\Pr(D)/S & \mbox{if $D$ is well-formed} \\ 0 &
\mbox{otherwise}\end{array} \right\}$$ 
Conveniently, since $S$ is a
constant, maximizing $\Pr_{Normalized}(D)$ is the same as maximizing
$\Pr(D)$ over just well-formed structures (or classes) $D$.

\subsection{Similarities among the models}

The general form of models A, B, and C is motivated in
\cite{eisnercoling}, to which the reader is referred for discussion.
Figure~\ref{fig:modelnarrative} also gives an overview.

When defining precise versions of the models, we took care to make the
comparison fair, by having the models condition their decisions on
comparable information.  In particular, all the models are sensitive
to the following types of probabilistic interactions:
\begin{enumerate}
\item\label{kidseq} {\em Words are conditioned on their parents and on adjacent siblings in the dependency tree.}
Each child is attached (model A) or predicted (model B) or
generated (model C) with a probability that varies with its
$\angles{\mbox{word}, \mbox{tag}}$ pair and the $\angles{\mbox{word},
\mbox{tag}}$ pair of its parent.  The probability also varies with the
tag of the next-closest child of that parent, so that the model is
sensitive to Markov-like dependencies among the successive
children of a word.
\item\label{stringseq} {\em Words are conditioned on adjacent words in the string.}
Except in model C, which does not attend to string-local
relations, each $\angles{\mbox{word}, \mbox{tag}}$ pair is generated with probability
conditional on the two $\angles{word,tag}$ pairs that immediately
precede it in the string.
\item When generating a sequence of $\angles{\mbox{word}, \mbox{tag}}$
pairs, we begin in a distinguished start state, and end when we
generate a distinguished ``stop'' symbol.  (We generate such sequences
in items~\ref{kidseq}--\ref{stringseq} above: the sequence of children of a word
is essentially regarded as the output of a Markov process, as is the sequence
of words in the sentence.)
\item Backoff and smoothing for sparse data (see \S\ref{sec:smooth})
are performed similarly for all models.  For example, in any model where
the probability of a child depends on the identity of its neighboring sibling, 
we consider only a shortened version of that sibling's tag.
\end{enumerate}

\begin{figure}
\rule{\textwidth}{3pt}
\begin{minipage}{\textwidth}
How does {\em play} in Figure~\ref{sampleparse} receive its two left
children ({\em dachshund}, {\em to})?

\begin{description}
\item[Model A:] {\em play} considers all its predecessors from nearest to
farthest: {\em to}, {\em dachshund}, {\em his}, {\em taught}, {\em
corner} \ldots.  It decides yes, yes, no, no, no, \ldots  

This sequence of decisions has a probability, like a sequence of coin
flips, but the flips here are not independent: each decision may be
influenced by the word selected on the most recent ``yes.''  In
principle, when {\em to} attaches to {\em play}, it removes the need
for subject-verb agreement, thereby increasing the chance that the
singular noun {\em dachshund} will attach.  And when {\em dachshund}
attaches, it fills the subject position of play, thereby reduces the
chance that another noun {\em man} will attach from the left.

The total probability of a possible dependency structure is the
probability of generating the words and tags in the structure by a
Markov process, times the probability of achieving exactly the right
sequence of $(n+1) \cdot n$ coin flips to obtain the observed links.

\item[Model B:] {\em play} decides at random that it wants {\em to} as
its closest left child.  Based on this, it decides that it wants {\em
dachshund} as its next closest child, and no more children beyond
that.  By a fabulous coincidence, it happens that {\em to} and {\em
dachshund} have just been generated by a Markov process.  They are
used to fill {\em play}'s requirements.\footnote{This narrative
ignores the role of {\em to}, which may in the same spirit be hoping
for an instance of {\em play} to serve as its parent.}

The total probability of a possible dependency structure is the
probability of generating the words and tags in the structure by a
Markov process, times the probability that each word {\em a
priori} wants children and a parent like those it is assigned in the
structure.

\item[Model C:] {\em play} decides at random that it wants {\em to}
and {\em dachshund}, and nothing else, as its left children.  This
choice is made by a Markov process exactly as in Model B, but here as
a result of {\em play}'s decision, the desired children are actually
generated; there is {\em not} a separate Markov process to generate
the words and tags.

The total probability of a possible dependency structure is the probability
that each word would {\em a priori} want children like those that it
has in the structure.

\item[Model D:] Just like model B,\footnote{Without parent
preferences.} except that the probability that {\em play} chooses {\em
dachshund} is conditioned on the fact that {\em dachshund} is 
available to serve as subject, i.e., was previously generated
in an appropriate position by a Markov process.

To see the difference, imagine a society where dachshunds are very
rare, but where they are sterotypically playful.  In model B, {\em
play} might reject {\em dachshund} as a subject, on the grounds that
{\em man} is a more common subject for {\em play}.  In model D, however,
{\em play} would be quite eager to accept {\em dachshund} as a subject,
because when {\em dachshund} is in the same sentence as {\em play}, in 
a position where it can serve as subject, it is very likely to do so.

The total probability of a possible dependency structure is the
probability of generating the words and tags in the structure by a
Markov process, times the probability that each word would select
the children it has in the structure, given that they are available
to be selected.  This is the same probability that model A would
compute if model A only multiplied the probability of the Markov
process and the $n$ ``yes'' coin flips while ignoring the $n^2$
``no'' coin flips.
\end{description}
\end{minipage}
\rule{\textwidth}{3pt}
\caption{Understanding the models by example.}\label{fig:modelnarrative}
\end{figure}

\subsection{Description of the individual models}\label{sec:indivmodels}

The pseudocode given below is never actually run, but merely
illustrates how, according to each model, a dependency structure $D$
with probability $\Pr(D)$ would be generated via a canonical series of
probabilistic choices.  

It is straightforward to reconstruct from the pseudocode how $\Pr(D)$
is defined for a given structure $D$.  It is this value that the
parser maximizes through dynamic programming, as described in \cite{eisnercoling}.

{\bf Notation:} $tw_i$ denotes the pair $\angles{w_i, t_i}$, called a
``tagged word.''  The parents $p_i$ are represented as indices, so
that $p_i = j$ means that the $i$th word modifies the $j$th word.  The
indices of the closest, 2nd-closest, 3rd-closest \ldots right children
of $w_i$ are sometimes denoted by $kid(i,1), kid(i,2), kid(i,3),
\ldots$, and similarly the indices of the left children are
$kid(i,-1), kid(i,-2), \ldots$.  Also, we let $kid(i,0)$ denote $i$
itself, but by abuse of notation, $tw_{kid(i,0)}$ is taken to
represent not $tw_i$ but a distinguished value $\angles{\mbox{\sc
bokids}, \mbox{\sc bokids}}$ that indicates the beginning of the left
or right child sequence.

\begin{description}
\item[Model A: Bigram lexical affinities] Model A first generates a
tagged sentence, using a simple trigram Markov model, as follows:
\begin{prog}
\progline $\Pr(D) := 1$
\progline $tw_{-1} := tw_0 := \angles{\mbox{\sc bos}, \mbox{\sc bos}}$ \scomment{beginning-of-sentence} 
\progline n := 0
\progline {\bf for} $n$ {\bf from} 0
\progline \>choose $tw_{n+1}$ randomly from among all possible tagged words, conditioned on $tw_{n-1}$ and $tw_{n}$
\progline \>$\Pr(D) := \Pr(D) \times \Pr(tw_{n+1} \mid tw_{n-1}, tw_n)$
\progline \>{\bf if} $tw_{n+1} = \angles{\mbox{\sc eos}, \mbox{\sc eos}}$ {\bf then break} \scomment{end of sentence; don't change $n$}
\end{prog}

At this point, the model has generated a sequence of tagged words that
has some probability $\Pr(\vec{w}, \vec{t})$.  In the second phase,
the model will choose parents conditional on these tagged words, to
get a full dependency structure with probability $\Pr(\vec{p}
\mid \vec{w}, \vec{t}) \times \Pr(\vec{w}, \vec{t}) = \Pr(\vec{w},
\vec{t}, \vec{p}) = \Pr(D)$.

For each pair of words (including the {\sc eos} mark), the second
phase of the model chooses whether to add a link between them.  The order in which
it makes these decisions is important, since links are not chosen
independently.  When $w_i$ decides whether to become a child of $w_k$,
it can condition its decision on the next-closest child of $w_k$.
This lets the model capture certain facts about subcategorization.
\begin{prog}
\progline {\bf for} $k := 1$ {\bf to} $n+1$  
\progline \>\comment{choose the left children of $w_k$: each choice sees $w_k$ and the tag of the next-closest child}
\progline \>\>c := 0 \scomment{number of left children so far}
\lprogline{modelA:line:chooseleftkids}
          \>\>{\bf for} $i := k-1$ {\bf downto} $1$
\progline \>\>\>choose whether $k$ should take $i$ as $kid(k,-(c+1))$, conditioned on $tw_i$, $tw_k$, and $tw_{kid(k,-c)}$   
\lprogline{modelA:line:linkprob}
          \>\>\>$\Pr(D) := \Pr(D) \times \Pr(\mbox{the above yes/no choice} \mid tw_i, tw_k, tw_{kid(k,-c)})$
\progline \>\>\>{\bf if} we chose ``yes'' 
\progline \>\>\>\>{\bf then} $c := c+1$ and $kid(k,-c) := i$
\progline \>\comment{likewise choose the right children of $w_k$}
\progline \>\>c := 0 \scomment{number of right children so far}
\progline \>\>{\bf for} $i := k+1$ {\bf to} $n$
\progline \>\>\>$\vdots$\scomment{as above}
\end{prog}

This model is leaky because many of the links are chosen independently
of each other, so that it is possible to generate illegal dependency
structures that feature words with no parents, words with multiple
parents, link cycles, or crossing links.

\item[Model B: Selectional preferences] The first phase of model B
generates a tagged sentence, exactly as in model A.  Thus each word
and its tag are chosen based on local context.  

The second phase corresponds to the pseudocode below.  For each word
it randomly and independently chooses a highly specific
subcategorization/supercategorization frame from among all such
frames, and then tries to link the words together so as to satisfy all
the frames at once.  The frame of a given word describes the parent
and children that the word expects; it corresponds to a lexicalized
version of a ``disjunct'' in link grammar \cite{linkgrammar} or a
``supertag'' in probabilistic tree-adjoining grammar \cite{supertags}.  

A tagged word $tw$ chooses the subcategorization part of its frame by
generating a Markov sequence of desired left children and another
Markov sequence of desired right children, that is, of tagged words
that $tw$ expects to match the heads of its complements and adjuncts.
It gets the supercategorization part of its frame by choosing a tagged
word that it expects to match its parent.  All generation
probabilities are further conditioned on $tw$ itself.
\begin{prog}
\progline {\bf for} $k := 1$ {\bf to} $n+1$  
\progline \>\comment{choose a sequence of tagged words that $tw_k$ will require its left children to match}
\nprogline\>\comment{each choice sees $tw_k$ and the tag of the next closest child}
\progline \>\>{\bf for} $c$ {\bf from} $0$
\lprogline{modelB:line:subcatchoice}
          \>\>\>choose a tagged word $\hat{tw}(k,-(c+1))$ that $kid(k,-(c+1))$ will have to match;
\nprogline\>\>\>\>choice is made over all tagged words in the vocabulary, conditioned on $tw_k$ and $\hat{tw}(k,-c)$
\lprogline{modelB:line:subcatchoiceprob}
          \>\>\>$\Pr(D) := \Pr(D) \times \Pr(\hat{tw}(k,-(c+1)) \mid tw_k, \hat{tw}(k,-c))$ 
\progline \>\>\>{\bf if} $\hat{tw}(k,-(c+1)) = \angles{\mbox{\sc eokids}, \mbox{\sc eokids}}$ {\bf then break} \scomment{end of left child sequence}
\progline \>\comment{similarly choose a sequence of tagged words that $w_k$ will require its right children to match}
\progline \>\>$\vdots$\scomment{as above}
\progline \>{\bf if} $k \leq n$
\progline \>\>choose a tagged word $\hat{tw}$ that $w_k$ will require its parent to match, conditioned on $w_k$
\progline \>\>$\Pr(D) := \Pr(D) \times \Pr(\hat{tw} \mid w_k)$ 
\lprogline{modelB:line:resolve} 
          If possible, choose parents $p_1,p_2,\ldots p_n$ in any way so that the parent and children of each word 
\nprogline\>$w_k$ satisfy the frame for $w_k$ generated above.
\end{prog}

This model is leaky: step~\ref{modelB:line:resolve} may fail, because
the right words to satisfy the frames are not present, or else cannot
be linked together without crossing links.  In that case, some
probability mass has been assigned to an impossible structure.  Note
in particular that each tagged word is generated two or more
times---once during the generation of a tagged sentence and once for
each of the other words it links to as parent or child.  The structure
can be legal only if the same tagged word is generated on all these
occasions.  Furthermore, the model is incomplete, because there may be
more than one way to carry out step~\ref{modelB:line:resolve}.%
\footnote{\label{fn:modelBincomplete} For example, if we were parsing arithmetic expressions, the
9-word string $a-x-p-q-y$ (not a minimal example) would admit both of
the following dependency strucures, which correspond to
$a-(x-((p-q)-y))$ and $(a-((x-(p-q))-y))$, and which reflect exactly
the same frame for each word token $w_i$:
$$\Tree [ a [ x [ [ p q ].- y ].- ].- ].- \hspace{1in} \Tree [ a [ [ x
[ p q ].- ].- y ].- ].- $$ 

It can be proved that if each word's frame specifies not only the
number of its left and right children, but also the direction (left or
right) of its parent, then at most one dependency structure is
possible in step~\ref{modelB:line:resolve} and the model is no longer
incomplete.  Unfortunately, \S\ref{sec:results} shows that augmenting 
frames with parental direction hurts the model's performance substantially.}

\item[Model C: Recursive generation] Model C is based on the idea that
each word generates its {\em actual} children, in just the same way
that in Model B each word generates its {\em desired} children
(subcategorization frame).  Given a tagged word $tw$, we can generate
its left children $tw(-1), tw(-2), \ldots$ as a Markov sequence of
tagged words further conditioned on $tw$, and its right children
likewise.\footnote{We use this notation for convenience, rather than
describing how to work out the actual indices of the children, which
is straightforward.}  The process is repeated recursively on those
words, yielding a tree as in Figure~\ref{sampleparse}b.  The process
consists of calling {\em generate} on the pair $\angles{\mbox{\sc
eos}, \mbox{\sc eos}}$, where {\em generate}$(tw)$ is defined below.
\begin{prog}
\progline \>\comment{choose $tw$'s sequence of left children, $tw(-1), tw(-2), \ldots$; each choice sees $tw$ and the tag of the next closest child}
\progline \>\>{\bf for} $c$ {\bf from} $0$
\progline \>\>\>choose $tw(-(c+1))$ conditional on $tw$ and $tw(-c)$
\progline \>\>\>$\Pr(D) := \Pr(D) \times \Pr(tw(-(c+1)) \mid tw, tw(-c))$ 
\progline \>\>\>{\bf if} $tw(-(c+1)) = \angles{\mbox{\sc eokids}, \mbox{\sc eokids}}$ {\bf then break} \scomment{end-of-left-child-sequence}
\progline \>\>\>{\em generate}$(tw(-(c+1)))$
\progline \>\comment{similarly choose a sequence of tagged words that $w_k$ will require its right children to match}
\progline \>\>$\vdots$\scomment{as above}
\end{prog}
This top-down, generative model is neither leaky nor
incomplete.\footnote{An interesting variation would be to do the
generation in depth-first order, so that the $c$th child of a word (on
either side of the word) would be chosen based on the word, its left
sibling, {\em and} the two tagged words that immediately string-precede the
child's subtree.  This would allow the model to take string-local context
into account in the same way as models A and B.  This variation has
not been tested experimentally.  While it can be implemented within the
parsing framework of \cite{eisnercoling}, it leads to somewhat larger span signatures, 
meaning that its time and space requirements suffer from a fairly
large (though not wholly unreasonable) constant factor.  On the other hand,
the model could consider {\em untagged} versions of the preceding words 
at no extra cost (Mike Collins, p.c.).}
\end{description} 

\noindent In addition to models A, B, and C, originally described in
\cite{eisnercoling}, we report results for a new bottom-up model that
is similar to both model A and model B, as well as to the model of
\cite{collins96}:

\begin{description}
\item[Model D: Realistic selectional preferences] It is simplest to
regard this model as a variant of model B.  When a word in model B
generates its subcategorization frame
(line~\ref{modelB:line:subcatchoice}), it does so in ignorance of the
words that are actually available to fill such a frame---although
those words have already been generated.  A better model would
generate a string of words as model B does, and then have each word
select a sequence of real children (tokens) from among the remaining
words (cf.\ model A) rather than generating a sequence of desired
children (types).  Ideally, the latter phase of this model would look
as follows:
\begin{prog}
\progline {\bf for} $k := 1$ {\bf to} $n+1$  
\progline \>\comment{select the left-child sequence of word $k$ from among existing words}
\progline \>\>{\bf for} $c$ {\bf from} $0$
\progline \>\>\>choose $kid(k,-(c+1))$ from the set $C = \{1, 2, \ldots kid(k,-c)-1, {\sc eokids}\}$
\nprogline\>\>\>\>i.e., the choices are the words to the left of $kid(k,-c)$ plus the distinguished symbol {\sc eokids}
\lprogline{modelD:line:hardprob}
          \>\>\>$\Pr(D) := \Pr(D) \times \Pr(kid(k,-(c+1)) \mid C, tw_k, tw_{kid(k,-c)})$ 
\progline \>\>\>{\bf if} $kid(k,-(c+1)) = {\sc eokids}$ {\bf then break} \scomment{end of left child sequence}
\progline \>\comment{similarly select the right-child sequence of word $k$}
\progline \>\>$\vdots$\scomment{as above}
\progline \>\comment{no parent or supercategorization frame is chosen}
\end{prog}
This model does generate structures with crossing links, so it is 
leaky, but less so than model B. 

The difficulty with the description above lies in estimating
$\Pr(kid(k,-(c+1)) \mid C, tw_k, tw_{kid(k,-c)})$ in
line~\ref{modelD:line:hardprob}.  This is the probability of choosing
a particular next child $i$ (perhaps the special choice {\sc eokids})
given the particular set $C$ of available remaining children.

The model of \cite{collins96} faces a similar problem, which Collins
addresses essentially by using the backed-off probability
$\Pr(kid(k,-(c+1))=i \mid tw_i, tw_k)$.  That is, when tokens labeled
like $tw_i$ and $tw_k$ are in the sentence, what is the probability
that the former links to the latter?  

Model D also backs off, but in a more nuanced way, to
$\Pr(kid(k,-(c+1))=i \mid tw_i, (i \in C), tw_k, tw_{kid(k,-c)})$.
That is, when tokens labeled like $tw_i$ and $tw_k$ are in the
sentence, what is the probability that the former links to the latter
as the $(c+1)$st left child---given that the former does fall to the
{\em left} of the latter's $c$th left child, which is labeled like
$tw_{kid(k,-c)}$?  These last two conditions are disregarded by 
\cite{collins96}.

Note that the more nuanced backoff of Model D enables the model to
capture probabilistic interactions among successive children of $w_k$,
just as models A--C do.  The most important such interaction is that
between the outermost child of $w_k$ and {\sc eokids}.  This
interaction serves to capture such facts as arity---the fact that
$w_k$, depending on whether it has an existing child of a particular
type, may require or forbid an additional child.

For purposes of computation, one might regard model D as a variant of
model A.  The difference (other than the use of {\sc eokids}) is that
in line~\ref{modelA:line:linkprob} of model A, we will continue to
multiply $\Pr(D)$ by the probabilities that the links we have accepted
are indeed right, but {\em not} by the probabilities that the links we
have rejected are indeed wrong.  Thus, the probability that model D
assigns to a structure is the probability that model A would select
{\em at least} the $n$ links found in the structure, and perhaps
others as well.  It is difficult to give an independent justification
of the model along these lines: note especially the problem that if
model A assigns some probability to a structure with $m > n$ links,
then this probability is added separately to {\em each} of the
\chooz{m}{n} structures with some $n$ of these links.
\end{description}

\section{Probability estimation}\label{sec:smooth}

The conditional probabilities required by \S\ref{sec:models} are difficult
to estimate directly, as they represent ratios of counts of rare
events that may never have occurred in training data.  The present
section describes how we estimate the probabilities.

\subsection{Overall backoff strategy}

The general approach is to decompose a probability of the form
$$\Pr(A, B, C \mid D, E)$$ into a product of the form $$\Pr(C \mid D, E)
\times \Pr(B \mid C,D,E) \times \Pr(A \mid B,C,D,E)$$ Each of the {\bf
factors} in this product is then estimated separately as a conditional
probability.

To estimate a conditional probability, we may have to back off, i.e.,
{\bf reduce} the condition so that it is less detailed.  For example,
to estimate the third factor above, we might choose to reduce its
condition $B, C, D, E$ to simply $B$, or perhaps to $D, E$, depending
on the independence assumptions that we believe are justified.  That
is, we might estimate $\Pr(A \mid B,C,D,E)$ as $\Pr(A \mid B)$ or
$\Pr(A \mid D,E)$ respectively.

Severe reductions throw away much potentially relevant information
about the conditions that obtain in the sentence, so they are
justified only by sparse data.  To allow a dynamic tradeoff between
sensitive conditions and sufficient data, each conditional probability
is estimated using an associated {\em list} of reductions, which are
increasingly more severe.  The first reduction on the list keeps all
or most of the original condition; later reductions throw away more
and more of this information.

How is this list of reductions used?  Suppose that we wish to estimate
$\Pr(A \mid B,C,D,E)$, and that the first (least severe) reduction
reduces the condition to $D, E$.  If this is the only reduction, we
return the estimate $\frac{count(A \& D \& E)+0.005}{count(D \&
E)+0.5}$, an approximation to $\Pr(A \mid D,E)$ that is based on
training data counts.  However, if there are additional reductions, we
recursively compute an estimate $p$ using the remaining list of
reductions, and return the estimate $\frac{count(A \& D \& E) +
3p}{count(D \& E) + 3}$.  Thus, the coarse, backed-off estimate $p$
has the weight of 3 additional observations of the specific context
$D,E$.  If the specific context $D,E$ has been frequently observed, it
will largely override the coarse estimate $p$
\cite{collins96}.\footnote{For efficiency, we do not bother to add
$3p$ to the numerator and 3 to the denominator, or even to compute $p$,
in the event that $count(D \& E)$ is large $\geq 8$.  This policy may
actually be unwise, given that the numerator may be zero even when
the denominator is large.}

\subsection{Features and functions used for backoff}\label{sec:features}

In describing the factors and reductions, we assume the following
functions:
\begin{description}
\item[$tag(tw)$:] Given a tagged word, extract the tag.\footnote{While
it is not done here, the $tag$ function can be modified so that if
$tw$ is a ``special'' word, such as a high-frequency or closed-class
word, then $tag(tw) = tw$.  This is equivalent to giving special words
their own tags.}
\item[$word(tw)$:] Given a tagged word, extract a lowercase version of the word.
\item[$cap(tw)$:] Given a tagged word, extract information about its capitalization.
$cap(tw)$ can take on four values: 
   \begin{description}
   \item[DOWN] all-lowercase
   \item[UP] all-caps (and at least two letters)
   \item[INIT] $tw$ is the first non-punctuation word in a sentence, and just its first letter is capitalized.
   \item[CAP] all other cases, in particular unambiguously capitalized or mixed-case words
   \end{description}
\item[$short(tag)$:] A function that maps specific tags to more general tags: for example, 
   $short(\verb+JJR+) = short(\verb+JJS+) = \verb+ADJ+$.  The corpus we use has 71 tags but 
   only 22 shortened tags.
\item[$tiny(tag)$:] A more aggressive version of $short(tag)$, which groups the 71 tags into just
   7 equivalence classes: Noun, Verb, Noun Modifier, Adverb, Preposition, Wh-Word, and Punctuation.
\item[$dist(i,j)$:] The distance between word positions $i$ and $j$, represented as one of
    the four ranges 1, 2, 3--6, and 7--$\infty$.
\end{description}

\subsection{Reductions used for backoff in models A, B, C, and D}

We now turn to the probabilities that must be generated.
\begin{enumerate}
\item Models A, B, and D must each generate a string of tagged words 
according to a trigram model.  The crucial probability is
computed via the following factors and reductions:
\begin{eqnarray*}
\Pr(tw_{k+1} \mid tw_{k}, tw_{k-1})  
 & = & \Pr(cap(tw_{k+1}), word(tw_{k+1}), tag(tw_{k+1}) \mid tw_{k}, tw_{k-1}) \\
 & = & \Pr(tag(tw_{k+1}) \mid tw_{k}, tw_{k-1}) \\
 &   & \hspace{0.45in}\mbox{reduction list}:
       \begin{array}{|l|}\hline
          tag(tw_{k}), tag(tw_{k-1}) \\ \hline
          tag(tw_{k})                \\ \hline
          short(tag(tw_{k}))         \\ \hline
       \end{array} \\
 &   & \times \Pr(word(tw_{k+1}) \mid tag(tw_{k+1}), tw_{k}, tw_{k-1}) \\
 &   & \hspace{0.45in}\mbox{reduction list}: 
       \begin{array}{|l|}\hline
          tag(tw_{k+1}) \\ \hline
       \end{array} \\
 &   & \times \Pr(cap(tw_{k+1}) \mid word(tw_{k+1}), tag(tw_{k+1}), tw_{k}, tw_{k-1}) \\
 &   & \hspace{0.45in}\mbox{reduction list}:
       \begin{array}{|l|}\hline
              word(tw_{k+1}), tag(tw_{k+1}) \\ \hline            
              tag(tw_{k+1}) \\ \hline
       \end{array} \\
\end{eqnarray*}

Notice that if we consider only the first reduction for each factor,
the above specifies the estimate $\Pr(t_{k+1} \mid t_k, t_{k-1}) \times
\Pr(w_{k+1} \mid t_{k+1})$, which is a standard trigram model
\cite[and others]{Church:trigram}.

\item Model B must also estimate $\Pr(tw_{parent} \mid tw_{child})$.
This is formally the same problem as estimating $\Pr(tw_{k+1} \mid tw_k)$,
and is handled as a special case of the $\Pr(tw_{k+1} \mid tw_k, tw_{k-1})$ 
computation above.

\item Models B and C must generate $tw_k$'s sequence of children in
direction $dir$, where $dir$ is either Left or Right.  For concreteness
let us assume that $dir$=Right, so that the children are $kid(k,1), kid(k,2), \ldots$.
\begin{eqnarray*}
 \lefteqn{\Pr(tw_{kid(k,c+1)} \mid tw_k, tw_{kid(k,c)}, dir)} \\
 & = & \Pr(cap(tw_{kid(k,c+1)}),word(tw_{kid(k,c+1)}),tag(tw_{kid(k,c+1)}) \mid tw_k, tw_{kid(k,c)}, dir) \\
 & = & \Pr(tag(tw_{kid(k,c+1)}) \mid tw_k, tw_{kid(k,c)}, dir) \\
 &   & \hspace{0.45in}\mbox{reduction list}:
       \begin{array}{|l|}\hline
	      tw_k, short(tag(tw_{kid(k,c)}))), dir \\ \hline
	      \left\{ \begin{array}{l}
   	                tw_k, dir  \\  
                        tag(tw_k), short(tag(tw_{kid(k,c)})), dir \\
                     \end{array} \right. \\ \hline 
	      short(tw_k), dir \\ \hline
       \end{array} \ \ \ \mbox{(see below)} \\
 &   &\times \Pr(word(tw_{kid(k,c+1)}) \mid tag(tw_{kid(k,c+1)}), tw_k, tw_{kid(k,c)}, dir) \\
 &   & \hspace{0.45in}\mbox{reduction list}:
       \begin{array}{|l|}\hline
	      tag(tw_{kid(k,c+1)}), tw_k, dir    \\ \hline
	      tag(tw_{kid(k,c+1)}), tag(tw_k), dir \\ \hline
	      tag(tw_{kid(k,c+1)}) \\ \hline
       \end{array} \\
 &   & \times \Pr(cap(tw_{kid(k,c+1)}) \mid word(tw_{kid(k,c+1)}), tag(tw_{kid(k,c+1)}), tw_k, tw_{kid(k,c)}, dir) \\
 &   & \hspace{0.45in}\mbox{reduction list}:
       \begin{array}{|l|}\hline
	      word(tw_{kid(k,c+1)}), tag(tw_{kid(k,c+1)}) \\ \hline
	      tag(tw_{kid(k,c+1)}) \\ \hline
       \end{array} \\
\end{eqnarray*}

In a version of model B or C in which we generate not only the
children of a head but also their desired distance from the head, the
objective is to find $\Pr(dist(k,kid(k,c+1)), tw_{kid(k,c+1)} \mid
tw_k, tw_{kid(k,c)}, dir)$.  For this version we must multiply by an
additional factor:
\begin{eqnarray*}
 &   & \times \Pr(dist(k,kid(k,c+1)) \mid tw_{kid(k,c+1)}, tw_k, tw_{kid(k,c)}, dir) \\
 &   & \hspace{0.45in}\mbox{reduction list}:
       \begin{array}{|l|}\hline
	      tw_{kid(k,c+1)}, tag(tw_k) \\ \hline
	      tag(tw_{kid(k,c+1)}), tag(tw_k) \\ \hline
       \end{array} \\
\end{eqnarray*}
Note that while model C is ordinarily not a leaky model, adding this
factor will make it leaky.

Note that one of the reductions above has a disjunctive condition,
whose {\bf disjuncts} are grouped by a bracket \{.  Disjunction is
useful because it is not always possible to know which part of the
original condition should be thrown away in a reduction in order to
overcome sparse data.  Using this disjunctive reduction, we estimate
the desired factor $\Pr(tag(tw_{kid(k,c+1)}) \mid tw_k, tw_{kid(k,c)},
dir)$ as follows:
   $$\frac{count(tag(tw_{kid(k,c+1)}), tw_k, dir) 
            + count(tag(tw_{kid(k,c+1)}), tag(tw_k), short(tag(tw_{kid(k,c)})), dir))}{count(tw_k,dir)+count(tag(tw_k), short(tag(tw_{kid(k,c)})), dir)}$$
That is, we compute a numerator and denominator for each 
disjunct separately, as if that disjunct were the entire reduction, and find our estimate by adding the
numerators and adding the denominators.
The disjunct with the greater denominator (i.e., whose condition
is more common) will have the greater influence on our estimate  \cite{collinsbrooks95}.

\item Models A and D must be able to decide, for words at positions
$k$ and $i > k$ such that $k$ already has $c$ children between $k$ and
$i$, whether $i$ is the $(c+1)$st child of $k$:
\begin{eqnarray*}
\lefteqn{\Pr(\mbox{link from $i$ to $k$} \mid tw_i, tw_k, tw_{kid(k,c)})} \\
 &   & \hspace{0.45in}\mbox{reduction list}:
       \begin{array}{|l|}\hline
	       word(tw_i), tag(tw_i), word(tw_k), tag(tw_k), short(tag(tw_{kid(k,c)})) \\ \hline
	      \left\{ \begin{array}{l}
	                   tag(tw_i), word(tw_k), tag(tw_k), short(tag(tw_{kid(k,c)})) \\
	       word(tw_i), tag(tw_i),             tag(tw_k), short(tag(tw_{kid(k,c)})) \\
	       word(tw_i), tag(tw_i), word(tw_k), tag(tw_k) \\
	      \end{array} \right. \\ \hline 
	                tag(tw_i),             tag(tw_k), short(tag(tw_{kid(k,c)})) \\ \hline
	                tag(tw_i),             tag(tw_k), tiny(tag(tw_{kid(k,c)})) \\ \hline
       \end{array} \\
\end{eqnarray*}

It is also possible to construct a version of these models that conditions
on distance, in which case we use the following:
\begin{eqnarray*}
\lefteqn{\Pr(\mbox{link from $i$ to $k$} \mid dist(i,k), tw_i, tw_k, tw_{kid(k,c)})} \\
 &   & \hspace{0.45in}\mbox{reduction list}:
       \begin{array}{|l|}\hline
	       dist(i,k), word(tw_i), tag(tw_i), word(tw_k), tag(tw_k), short(tag(tw_{kid(k,c)})) \\ \hline
	      \left\{ \begin{array}{l}
	                   dist(i,k), tag(tw_i), word(tw_k), tag(tw_k), short(tag(tw_{kid(k,c)})) \\
	       dist(i,k), word(tw_i), tag(tw_i),             tag(tw_k), short(tag(tw_{kid(k,c)})) \\
	       dist(i,k), word(tw_i), tag(tw_i), word(tw_k), tag(tw_k) \\
	      \end{array} \right. \\ \hline 
	      \left\{ \begin{array}{l}
	                   tag(tw_i), word(tw_k), tag(tw_k) \\ 
	       word(tw_i), tag(tw_i),             tag(tw_k) \\
	      \end{array} \right. \\ \hline 
	                dist(i,k), tag(tw_i),             tag(tw_k), short(tag(tw_{kid(k,c)})) \\ \hline
	                dist(i,k), tag(tw_i),             tag(tw_k), tiny(tag(tw_{kid(k,c)})) \\ \hline
       \end{array} \\
\end{eqnarray*}

\end{enumerate}

\subsection{Unknown words}

The system deals with unknown words in a uniform way, using a
technique of {\bf attenuation}.  Before parsing on a test sentence
begins, each unknown word in the input is {\bf attenuated}, that is,
replaced by a symbol indicative of the word's morphological class.
If the word ends in a digit, the symbol is \verb+MORPH-NUM+; else, if
it is 6 characters or more, the symbol is \verb+MORPH-+{\em XX}, where
{\em XX} are uppercase versions of the word's last two characters;
else the word is fairly likely to be monomorphemic and the symbol
\verb+MORPH-SHORT+ is used.  The capitalization properties of the
original word (see \S\ref{sec:features} above) are retained.

Formally, suppose $\vec{w}$ is the input word string and $A(\vec{w})$
is an attenuated version of the string in which unknown words have
been replaced with their morphological classes, as above.  Because the
parser is run on $A(\vec{w})$ rather than on $w$, it chooses tags
$\vec{t}$ and parents $\vec{p}$ so as to maximize $\Pr(A(\vec{w}),
\vec{t}, \vec{p})$.  But since $\Pr(\vec{w} \mid A(\vec{w}))$ is constant
given the input, this is the same as maximizing
$$\Pr(\vec{w} \mid A(\vec{w})) \cdot \Pr(A(\vec{w}), \vec{t}, \vec{p}) 
  \approx \Pr(\vec{w} \mid A(\vec{w}), \vec{t}, \vec{p}) \cdot \Pr(A(\vec{w}), \vec{t}, \vec{p}) 
  = \Pr(\vec{w}, \vec{t}, \vec{p})$$ 
as originally desired.

It is important that the system train on the attenuated symbols, such
as \verb+MORPH-+{\em XX}, and that the distribution of these symbols
during training correspond to the distribution of {\em unknown} words
(rather than all words) during testing.  The corpus we use (see
\S\ref{sec:corpus}) happens to be divided into discourse-coherent
sections (articles).  When training, we replace each word with its
morphological symbol throughout the entire first training section in
which it appears.%
\footnote{This strategy is computationally cheaper than the ideal
solution, which is to mine each sentence for statistics both on the
individual words {\em and} their morphological classes.  To avoid
sparse data problems, a training word is not replaced if it happens to
appear in test data.  In particular, if a word occurs only once in
training data, we will be careful to train on the full lexical item
and not merely its attenuation, if the full item will be needed for
parsing test data.  (This policy does not constitute ``peeking at the
test data when fitting the model,'' any more than case-based learning
does when it rescans the training data each time it needs to model a
test example.  It looks only at the input data, not the answer.)}

Thus, some of the sentences we train on include attenuated ``words''
such as \verb+MORPH-SHORT+.  The system is thereby able to learn, for
example, that tokens of unknown, short, lowercase words---i.e., short
lowercase words appearing in an article for the first time---tend to
be common nouns.  (By contrast, arbitrary tokens of short, lowercase
words are most often prepositions.)

\section{Description of the corpus}\label{sec:corpus}

Our corpus of dependency structures was derived from the {\em Wall
Street Journal} sentences that appear in the Penn Treebank II
\cite{TreebankII}.  For simplicity, we omitted sentences that contained
conjunction.  This allowed us to postpone questions about how best to
handle conjunction (either in constructing dependency representations
or in modifying the probability models).  We also omitted a number of
sentences in which we noticed clear annotator errors.  

Our corpus contained all 25,608 remaining sentences, whose lengths
ranged from 1 to 79 words including punctuation (mean 19, median 19).
The corpus was structured as 2,235 articles or sections of 1--130
sentences each (mean 11, median 6).

Each phrase-structure tree in the Penn Treebank was converted to a
bare-bones dependency structure (Figure~\ref{sampleparse}) by a
process of several steps:
\begin{enumerate}
\item Unflatten some instances of Treebank-style flat structure:
      \begin{itemize} 
      \item Group any maximal sequence of \verb+NNP+ (proper
      noun) siblings into a \verb+NPR+ (proper noun phrase)
      constituent.  

      \item Group any maximal sequence of \verb+CD+ (cardinal number)
      siblings into a \verb+QP+ (quantifier phrase).

      \item Group \verb+$ QP+ into a \verb+QPMONEY+ constituent.  
 
      \item Following each \verb+NPR+, group the maximal sequence of
      \verb+NN+ (common noun) siblings into a \verb+NP+ (noun phrase).
      \end{itemize}

\item Automatically correct a few common types of annotator errors, or
      discard the sentence if a correction cannot be effected automatically.

\item For each constituent $X$, from the bottom of the tree upward,
      use heuristics and exception tables to determine which of its
      overt (non-trace) subconstituents $Y$ contributes the head to
      $X$.  Define $head(X)$ to be the same as $head(Y)$.  For each
      overt subconstituent $Z \neq Y$, link $head(Z)$ to $head(Y)$, so
      that $head(Y)$ is the parent.

\item\label{tagset} Modify certain tags in the resulting structure to make them more informative:
      \begin{itemize}
	\item Mark auxiliary verbs as such (by adding a feature to their tags).

        \item Since premodifiers of nouns lose their ability to
              take complements or specifiers, mark them as such.

        \item Since participial postmodifiers of nouns lose their
              ability to take subjects, mark them as such.

        \item Distinguish complementizers from prepositions.  (In the Treebank,
              they share the tag \verb+IN+.)
      \end{itemize}

\item Modify the dependency structure so that it better reflects semantic relations:
      In a sequence of preverbal auxiliaries (possibly interrupted by
      adverbs), make each point to the next, and let the main verb
      be the head of the verb phrase.
\end{enumerate}

\section{Experiments and discussion}\label{sec:results}

\subsection{Evaluation method}

We divided the corpus randomly into test data (400 sentences) and
training data.  To deny ourselves the advantage of training on half an
article and testing on the other half, we chose the test data by
repeatedly choosing a sentence at random and marking its entire
section as test data, until we had marked 400 test sentences.  We 
scored the models on how well they tagged and parsed test data.

To evaluate the tagging, we found the percentage of words with the
correct tag.  Recall that we used a somewhat more fine-grained 
tag set (\S\ref{sec:corpus}, item~\ref{tagset}) than that used
by the Penn Treebank \cite{Treebank}, so the task was correspondingly harder.

To evaluate the parsing, we simply found the percentage of words that
attached correctly, i.e., that correctly selected their parents.  This
single {\bf attachment score} is easier to understand than a
(precision, recall, crossing brackets) triple as in PARSEVAL
\cite{parseval}.  As \cite{lin95} independently argues,\footnote{I am
grateful to Joshua Goodman for directing my attention to this paper.}
the attachment score also penalizes errors in a more appropriate way,
since under the PARSEVAL metrics, a single semantically difficult
misattachment can damage any number of constituents.

\subsection{Models evaluated}

As our current parser is written in Lisp, Model A was all but
impractical for us to run with a training set this large.  Recall that
model A has both high memory requirements (it must be able to remember
all pairs of words that have appeared in the same sentence) and high
time requirements (to compute the probability of even a known $n$-word
parse takes $O(n^2)$ time).  We terminated the experiment early, as
the test results on the early sentences appeared to be far inferior to
those of the other models.  

We ran the following versions of the other models.  Results are
shown in Figures~\ref{fig:results}--\ref{fig:fewerr}.

\begin{itemize}
\item A baseline model \cite{eisnercoling}.  Each word is tagged with
      the most common tag (ignoring case).  (Unknown words are treated
      in the usual way (\S\ref{sec:smooth}), so they are assigned the
      most common tag for new words sharing their capitalization and
      last two letters.)  Each tagged word now selects a parent: a word
      tagged $t$ will choose the parent at the most common distance
      for words tagged $t$.  For example, every determiner takes the
      following word as its parent.  The resulting structure may be
      ill-formed, but can still be scored on how many words had the
      correct tag and correct parent.
\item A trigram tagger, ``model X,'' that works identically to the
      first phase of models A, B, and D.  This tagger does not add any
      links; it is run so that we can compare its tagging accuracy to
      that of the parsers.
\item Three versions of model B, which vary in their attitudes toward supercategorization frames.
\begin{itemize}
\item The version of \S\ref{sec:models}, in which each word generates a desired parent.  
\item A version proposed in footnote~\ref{fn:modelBincomplete}, which
       remedies the incompleteness of model B by also having each
       word's supercategorization frame specify the direction in which
       the word's parent is to be found.\footnote{This introduces a new
       factor $\Pr(\mbox{parent dir} \mid \mbox{child},
       \mbox{parent})$ into the probability computation: we estimate
       it with the two reductions \fbox{$tag(\mbox{child}),
       tag(\mbox{parent})$} and \fbox{$short(tag(\mbox{child})),
       tiny(tag(\mbox{parent}))$}.}
\item A version in which no supercategorization frame is generated at all,
      so that line~\ref{modelB:line:resolve} in the pseudocode for model B has
      a higher chance of success.  
\end{itemize}

\item Three versions of model C were attempted:
\begin{itemize}
\item The pure version of \S\ref{sec:models}, in which each word generates sequences of
      left and right children.
\item A non-lexical version of the above, in which only the last (most severe) reduction
      is used for $\Pr(word(tw_{kid(k,c+1)}) \mid tag(tw_{kid(k,c+1)}), tw_k, tw_{kid(k,c)}, dir)$:
      this is estimated as $\Pr(word(tw_{kid(k,c+1)}) \mid tag(tw_{kid(k,c+1)}))$.  Thus the 
      statistical relation between a child and its parent is mediated only by the tags of the two words,
      and is ignorant of the words themselves.   This version corresponds to the straw-man model C$'$ of
      \cite{eisnercoling}.  
\item A leaky version that also generates a desired distance of each child from the head,
      as described in \S\ref{sec:smooth}.  This improves performance somewhat.
\end{itemize}

\item Model D, as described earlier in \S\ref{sec:models}.
\end{itemize}

\begin{figure}
\begin{tabular}{@{}r@{ }l|r||rrrrrr||r|}
\multicolumn{3}{l||}{}     & \multicolumn{6}{c||}{broken down by class of (correct) tag} \\                
   &                                  &  {\bf Non-punc} & N & V & NMod & Adv & Prep & Wh & Unknown \\ \hline
   & Word count (tokens)       & 7446 & 2067 & 1478 & 2555 & 303 & 958 & 85 & 248 \\ \hline
1) & Baseline                  & {\bf {\scriptsize 76.1 /} 41.9} &  29.9 &  34.5 &  51.8 &  32.7 &  56.4 &  34.1 & {\scriptsize 59.7 /} 32.7 \\
2) & Model X (trigram tagger)  & {\bf {\scriptsize 93.1 /} ------}&------& ------&  ------& ------&------&  ------ & {\scriptsize 82.7 /} ------ \\ \hline
3) & Model B, parent dir       & {\bf {\scriptsize 91.0 /} 71.5} &  67.3 &  64.7 &  90.7 &  70.6 &  44.4 &  20.0 & {\scriptsize 81.0 /} 70.2 \\
4) & Model B                   & {\bf {\scriptsize 92.8 /} 83.8} &  85.7 &  78.3 &  94.1 &  71.9 &  67.4 &  54.1 & {\scriptsize 82.3 /} 81.5 \\
5) & Model C, no lex\ \ (= C$'$)&{\bf {\scriptsize 92.8 /} 84.8} &  85.9 &  83.9 &  91.5 &  75.9 &  70.3 &  68.2 & {\scriptsize 81.6 /} 82.0 \\
6) & Model C                   & {\bf {\scriptsize 92.0 /} 86.9} &  88.6 &  83.2 &  91.9 &  79.2 &  80.0 &  69.4 & {\scriptsize 81.9 /} 86.7 \\
7) & Model C, distance         & {\bf {\scriptsize 92.0 /} 87.7} &  89.6 &  83.8 &  92.5 &  78.5 &  81.3 &  71.8 & {\scriptsize 82.3 /} 84.7 \\
8) & Model B, no supercat.     & {\bf {\scriptsize 93.7 /} 88.0} &  89.5 &  84.4 &  93.5 &  78.2 &  80.6 &  70.6 & {\scriptsize 83.1 /} 85.5 \\
9) & Model D                   & {\bf {\scriptsize 94.0 /} 90.0} &  91.3 &  87.5 &  95.0 &  83.8 &  80.9 &  74.1 & {\scriptsize 83.1 /} 87.5 \\ \hline
10) & Model C, distance, true-tags   & {\bf {\scriptsize ------  /} 90.4} &  92.8 &  87.9 &  95.1 &  80.2 & 81.7 &  70.6 & {\scriptsize ------  /} 90.7 \\
11) & Model D, true-tags        & {\bf {\scriptsize ------  /} 92.6} &  94.0 &  90.9 &  97.8 &  85.1 & 82.7 &  71.8 & {\scriptsize ------  /} 93.1 \\
12) & \cite{collins96}, auto-tags & {\bf {\scriptsize ------  /} 92.6} &  94.6 &  90.7 &  96.9 &  84.6 &  83.6 &  76.5 & {\scriptsize ------  /} 91.1 \\ \hline
\end{tabular}
\caption{Results for several models, in increasing order of
performance: how many words found their correct tag and parent?  Small
type shows the percentage of words whose {\em tags} were correctly
identified.  Large type shows the percentage of words whose {\em parents}
were correctly identified.  The first (boldfaced) column shows overall
scores, for all words other than punctuation.  The remaining columns
consider the models' performance on particular kinds of words, such as
prepositions, or unknown words (those not seen in the training
data).}\label{fig:results}
\end{figure}

\begin{figure}
\begin{tabular}{@{}r@{ }l|r||rrrrrr||r|}
\multicolumn{3}{l||}{}                   & \multicolumn{6}{c||}{class of true parent's (correct) tag} \\                
   &                          & Non-punc & N & V & NMod & Adv & Prep & Wh & Unknown \\ \hline
3) &Model B, parent dir       &  {\scriptsize 91.0 /} 71.5 & 72.9 & 67.8 & 73.3 &  51.0 & 62.5 & 59.3 & {\scriptsize 81.0 /} 70.2 \\
4) &Model B                   &  {\scriptsize 92.8 /} 83.8 & 84.0 & 77.6 & 81.2 &  69.4 & 91.5 & 84.9 & {\scriptsize 82.3 /} 81.5 \\
5) &Model C, no lex\ \ (= C$'$)& {\scriptsize 92.8 /} 84.8 & 85.9 & 82.3 & 74.5 &  51.0 & 86.6 & 84.9 & {\scriptsize 81.6 /} 82.0 \\
6) &Model C                   &  {\scriptsize 92.0 /} 86.9 & 86.9 & 83.7 & 84.7 &  71.4 & 91.1 & 80.2 & {\scriptsize 81.9 /} 86.7 \\
7) &Model C, distance         &  {\scriptsize 92.0 /} 87.7 & 88.2 & 84.3 & 84.5 &  69.4 & 91.8 & 81.4 & {\scriptsize 82.3 /} 84.7 \\
8) &Model B, no supercat.     &  {\scriptsize 93.7 /} 88.0 & 88.5 & 84.9 & 83.1 &  69.4 & 91.9 & 82.6 & {\scriptsize 83.1 /} 85.5 \\ 
9) &Model D                   &  {\scriptsize 94.0 /} 90.0 & 89.2 & 87.4 & 85.5 &  83.7 & 93.5 & 84.9 & {\scriptsize 83.1 /} 87.5 \\ \hline
10)&Model C, distance, true-tags   &  {\scriptsize ------  /} 90.4 & 90.7 & 86.7 & 84.9 &  69.4 & 94.7 & 86.0 & {\scriptsize ------  /} 90.7 \\
11)&Model D, true-tags             &  {\scriptsize ------  /} 92.6 & 92.1 & 90.1 & 84.9 &  79.6 & 95.9 & 87.2 & {\scriptsize ------  /} 93.1 \\
12)&\cite{collins96}, auto-tags      &  {\scriptsize ------  /} 92.6 & 94.0 & 91.0 & 86.7 &  85.4 & 95.6 & 94.0 & {\scriptsize ------  /} 91.1 \\ \hline
\end{tabular}
\caption{Essentially the same as in Figure~\protect{\ref{fig:results}}, except that
now the breakdown in the middle columns is different.  These columns
shows how well various parts of speech manage to ``recall'' their
children.  For example, what percentage of all words that should be
children of verbs are correctly attached to those
verbs?}\label{fig:parresults}
\end{figure}

\begin{figure}
\begin{tabular}{@{}r@{ }l|rrrrr||rrr||c|}
  &                            &  \multicolumn{5}{c||}{(a) Attachment errors} & \multicolumn{3}{c||}{(b) Contagion} & (c) Search\\
  &                            & 0   & $\leq 1$ & $\leq 2$ & $\leq 3$  & $\leq 4$ & p(err 1) & p(err 2) & Ratio & error \\ \hline
1)& Baseline                       &  0 &  3 &  4 &  6 & 11 & 100 & 97 & 0.97 &\\ 
2)& Model X (trigram tagger)       &    &    &    &    &    &     &    &      &16 \\ \hline
3)& Model B, parent dir            & 12 & 23 & 30 & 40 & 49 &  88 & 88 & 0.99 & 9 \\ 
4)&Model B                        & 28 & 43 & 54 & 62 & 72 &  72 & 79 & 1.10 & 8 \\ 
5)&Model C, no lex\ \ (= C$'$)    & 26 & 43 & 55 & 65 & 75 &  74 & 77 & 1.04 & 8 \\ 
6)&Model C                        & 32 & 50 & 62 & 72 & 78 &  68 & 74 & 1.08 & 14 \\  
7)&Model C, distance              & 33 & 48 & 64 & 72 & 82 &  67 & 78 & 1.16 & 16 \\ 
8)&Model B, no supercat.          & 34 & 52 & 66 & 75 & 82 &  66 & 73 & 1.10 & 17 \\ 
9)&Model D                        & 37 & 58 & 72 & 79 & 85 &  63 & 67 & 1.06 & 19 \\ \hline
10)&Model C, dist, true-tags   & 38 & 57 & 72 & 81 & 88 &  62 & 69 & 1.12 & 1 \\  
11)&Model D, true-tags             & 44 & 67 & 80 & 86 & 92 &  56 & 59 & 1.05 & 0 \\ 
12)&\cite{collins96}, auto-tags    & 47 & 66 & 79 & 86 & 93 &  53 & 64 & 1.21 & $\approx 0$ \\
   &                            &    &    &    &    &    &     &    &      & (Collins, p.c.) \\ \hline
\end{tabular}

\caption{{\bf (a)} Percentage of sentences with few attachment errors.  For
the better models, two-thirds of the sentences have 1 misattachment or
(usually) none at all.  Misattachments of punctuation are not counted.
{\bf (b)} When it rains it pours: Given that there is already one error in a
sentence, the probability of a second error is increased.  The columns
show $\Pr(\geq 1 \mbox{ err}$) and $\Pr(\geq 2 \mbox{ errs} \mid\;\geq
1 \mbox{ err}$), as percentages, as well as the ratio of these.  
{\bf (c)} The final column shows the percentage of sentences in each experiment
that are victims of search error.  For these sentences, the model
would have preferred the correct structure to the structure that the
parser found, but the parser did not consider it---perhaps because of
overly aggressive pruning in the parse chart, but typically because
the parser made wrong initial guesses about which tags to consider.
Indeed, providing tags to the parser (as in lines 10--11) essentially
eliminates search error, providing a clue to why performance improves
so much when tags are
provided.}\label{fig:fewerr}\label{fig:searcherr}
\end{figure}

\subsection{Results of the evaluation}

On the basis of our preliminary experiments \cite{eisnercoling}, we
expected model C to win.  Indeed model C did outperform models A
(apparently) and B.  However, it emerged that model C could be in turn
beaten by variations on model B, such as model
B-without-supercategorization (the third variant) and model D.  At
present, model D is our highest-performing model.

The non-lexical model C$'$ performs surprisingly well overall, only two percentage points
below the lexical version.  The two versions make rather
different errors.  The non-lexical version tends to favor
right-branching structure too strongly, whereas the lexical
version can be too easily led astray from right-branching
structure.  Better smoothing of low counts might help the latter
problem.  

Among the three variants of model B, the third was by far the most
successful.  Thus, supercategorization preferences appear to be
unreliable---as one might suspect from manually constructed competence
grammars, which traditionally focus on subcategorization.  For
example, it happens nouns more often modify words to their right
(including other nouns) than words to their left.  But this is
linguistically a fact not about nouns, but rather about the frequency
of words that wish to be modified by nouns from one side or another.
It is unwise to think that nouns insist on attaching rightward even at
the expense of subcategorization.

This third version of model B has an interesting property.  $\Pr(D)$
for a dependency structure $D$ happens to be exactly
$\Pr_{\mbox{\scriptsize model-C}}(D) \cdot \Pr_{\mbox{\scriptsize
model-X}}(D\mbox{'s tagged word sequence})$, where model X is a
trigram Markov model.  So the parser maximizes the product of the
generative probability (which considers only tree-local information)
and the Markov probability (which considers only string-local
information).  Compared to model C, which uses only the generative
probability, this hybrid does much better at tagging and slightly
better at parsing.  Compared to model X, which uses only the Markov
probability, the hybrid does slightly better at tagging and (of
course) much better at parsing.

Finally, a noteworthy result is that tagging can be improved by adding a
good parser, and vice-versa.  For the best models---model B without
supercategorization, and model D---tagging performance actualy beat
that of a pure trigram tagger, model X.  (For worse models, parsing
hurts tagging by overriding good decisions of the tagger.)  The
converse was also true: as just noted, model
B-without-supercategorization effectively beats model C by adding a
local tagger to it.

\subsection{Comparison with another parser}\label{sec:comparcollins}

In a further experiment, we compared the most successful versions of
the rather different models C and D to the state-of-the-art parser of
\cite{collins96}.  The results are shown on the last three lines of
Figure~\ref{fig:results}, and likewise for
Figures~\ref{fig:parresults}--\ref{fig:fewerr}.  Collins's parser
performs very similarly overall to our best model, model D.  There are
some fine-grained differences: for example, our model D has an
advantage on unknown words and nominal modifiers, while Collins's
parser is better at attaching (known) prepositions and wh-words.

For purposes of the experiment, the Collins parser was trained and
tested on the same Penn Treebank sentences that were presented to our
system.  We converted the output of the parser to dependency form
using the same automatic tools that we used to convert the Treebank
sentences (\S\ref{sec:corpus}).  This enabled us to score the output
using the same metrics.

One issue in making the comparison was that the Collins parser runs a
separate tagger, as a black box, before it begins to parse; this
tagger, unlike ours, is very highly trained.  To make the comparison fairer, we ran
our models in a mode where we informed them of the correct Treebank
tags.  This did not completely determine the more highly articulated
tags that our system actually uses (see \S\ref{sec:corpus}), but it
did constrain the choice of tags sufficiently to reduce tagging error
on our tag set to just 1.6\%.  (The tagger Collins uses
\cite{ratnaparkhi} has error of about 3\% on the Treebank tag set,
putting Collins at a slight disadvantage; but this is mitigated
somewhat by the fact that Collins trains his parser on the slightly
erroneous output of the tagger rather than the ``correct'' tags
(p.c.).)  The principal benefit of feeding tags to the model in this
way was that it virtually eliminated a quite serious problem of search
error (Figure~\ref{fig:searcherr}c), boosting our performance
substantially (Figure~\ref{fig:results}).

\subsection{Discussion of the comparison}\label{sec:comparcollinsdiscuss}

It is somewhat surprising that our accuracy roughly matches Collins's,
as our original plan \cite{eisnercoling} was to value simplicity
rather than high performance.  (We chose dependency grammar and
simple, homogeneous probability models because we wished to answer
some some key design questions about probabilistic parsing.)  

The two parsers use a number of mechanisms that are rather different,
and their probability models are mildly different in an interesting
way (see \S\ref{sec:indivmodels}).  Nonetheless, both parsers rely
heavily on associations between pairs of lexical items---and it is
possible to discern further points of correspondence:
\begin{enumerate}
\item The Collins parser has three parts, each of which uses a
different sort of probability model: tagging, chunking of ``base NPs,'' and
general parsing.  Our models are more homogeneous: they do not treat
base NPs specially, and model C does not even treat tagging specially.

Base NPs in the Collins parser carry roughly the same load as sibling
interactions in our parser.  In the Collins parser, base NPs help to
avoid certain errors: if the two words in ``John Smith'' or ``May
1996'' or ``water heater'' were not grouped into one object, the words
could both attach to a following verb.  Such ``double subject'' errors
do not arise in our parser, because the trick of generating children
as a terminated Markov sequence helps capture arity.  Verbs learn that
they should not have two nominal left children (in a row, at least).
Our parser's attention to the interdependencies among siblings
is in general an advantage, as the base-NP method will not capture
such interdependencies as the difference between transitive and
ditransitive verbs.

\item The Collins parser produces a tree labeled with nonterminal
symbols.  Because of these nonterminals, the probability model can
require that a verb have a subject if the resulting constituent is to
serve as a sentential complement.  (For example, {\em thought} takes a
sentential rather than a VP complement: in {\em I thought *(John) left (Mary),}
the verb {\em left} has low probability of linking to {\em thought} unless
it has a subject.)

In principle, the same work could be accomplished in a dependency
model by adding features to the tags, which would vary according to
the constituent structure ``projected'' by a head.  We have not
experimented with this.  However, all that is necessary is to allow
two tags for the verb {\em left}---one that has a high probability of
getting a subject, and the other that does not.  Only the former tag
is likely to link to {\em thought}.

Possibly a tree is more informative output than a bare-bones
dependency structure, because it may be easier to recover semantic
relationships given the additional internal structure.  However,
\cite{eisnercoling} notes that our methods could be easily adapted to
handle {\em labeled} dependencies rather than bare-bones dependencies.
That is, the links in Figure~\ref{sampleparse} could be annotated with
semantic roles or other symbols.  Phrase structure trees can be
more-or-less encoded with labeled dependencies of this sort (e.g., as
in \cite{collins96}), so the dependency methods described in
\cite{eisnercoling} are powerful enough to produce such trees.

\item The Collins parser is sensitive to intervening punctuation
between a parent and child (as well as other local configurations,
like adjacency and intervening verbs).  This is wiser than our
solution of treating punctuation marks as words, because it
recognizes that a single comma may discourage links at all levels
from crossing that comma.  
\end{enumerate}

Small differences of this kind in formalizing linguistic intuitions
can of course have substantial effects.  Indeed, such effects are the
topic of the current paper, and are amply demonstrated in
Figure~\ref{fig:results}.  However, it is apparently possible to make
the intuitions above work as intended.  In the best systems we have
considered here, including \cite{collins96} and our own model D and
most of the remaining errors are matters of semantic nuance, or more
precisely, nuances of semantic subcategorization that are difficult to
pick up from the syntax or from lexical associations seen in the
training data.

One would like to attack such errors without large sets of specialized
hacks or inordinate amounts of annotated training data.  It may help
to stop treating words atomically.  We may wish to explore methods
that can generalize beyond the affinities of individual words, and
classify words and phrases into groups, in terms of how they tend to
create and fill semantic roles.  Another approach would be to
bootstrap, i.e., to use the existing model to obtain approximate
parses for large quantities of naturally occurring raw text, which
could then be used for further training.

\subsection{Significance testing}

Particularly as the test sample consisted of only 400 sentences (7,446
words not counting punctuation), we wished to test our results for statistical
significance.  

It was necessary to be careful here, as attachment errors within the
same sentence might not be independent of each other.  Indeed,
Figure~\ref{fig:fewerr}b {\em shows} that they are not independent.
If parsing errors suffer from a ``when it rains it pours'' phenomenon
(i.e., the number of errors per sentence is a contagious
distribution), then an apparently large difference in
Figure~\ref{fig:results} might be the result of just one or two badly
parsed sentences with many attachment errors.  Thus, there is a danger
of finding significance where there is none.

For this reason we employed a non-parametric, Monte Carlo test.  Given
a pair of models whose error rates differed by $\mu$, we considered
the 400 pairs of parses produced by the models on the 400 test
sentences.  We asked: If for each pair of these parses we randomly
colored one parse red and the other parse blue, how often would the
difference between the red error rate and the blue error rate reach
$\mu$ or more?  (If the difference $\mu$ is due to one badly parsed
sentence, the answer is ``half the time''---not a good significance
level.)

For each pair of models in Figure~\ref{fig:results}, we
computed the significance level at which their attachment performance
differed, by making 10,000 random coloring passes through the 400
sentences and checking how often the random red-blue difference was as
strong as the observed difference between the models.

In the first column of Figure~\ref{fig:results}, nearly all of the
attachment performances were significantly different at the 0.005
level, and indeed usually at 0.001.  The only models that could not be
reliably distinguished were certain pairs on successive lines of
Figure~\ref{fig:results}.  To wit:
\begin{itemize}
\item adding distance to model C (lines 6 and 7) resulted in a significant improvement
   but only at the 0.05 level;

\item lines 4 and 5 were not significantly different, nor were lines 7 and 8,  9 and 10, or
      11 and 12.
\end{itemize}

We used the same technique to test the significance of differences in
tagging performance (for runs when the models were not given tags).
We found significance, again at the 0.005 level, in all but the
following comparisons (see Figure~\ref{fig:results}):
\begin{itemize}
\item lines 2, 4, and 5 were not significantly different from each other; that is,
    models B and C$'$ have the same tagging performance as a trigram tagger, model X;%
\footnote{It is interesting that C$'$ should have the same performance: it is 
     essentially identical to a trigram tagger, but where the trigrams consist
     of a parent and two of its adjacent children rather than three consecutive
     words.  In particular, the independence assumptions of both C$'$ nor X 
     allow words to influence on each other only through their tags.}
\item model C$'$ (line 5) does tag significantly better than the two versions of
    model C (lines 6--7), but only at the 0.05 and 0.07 levels respectively;
\item  the two versions of model C (i.e., with and without distance) do not significantly
    differ from each other;
\item the two best models (lines 8 and 9) do not significantly differ in tagging performance
  (though they do in parsing performance).
\end{itemize}

\section{Conclusions}\label{sec:conclusion}

We hope that this paper has been helpful in several ways.  First (and
foremost), the comparative results shed some light on how to design a
probability model for parsing.  In particular, for the models considered
here:
\begin{itemize}
\item lexical affinities are important (C vs. C$'$); 
\item it helps parsing to use string-local as well as tree-local information (B3 vs. C);
\item it helps tagging to use tree-local as well as string-local information (B3 vs. X);
\item it helps parsing to use distance information (C with and without distance);
\item it is harmful to assume that words generate supercategorization preferences (B3 vs. B1, B2) ;
\item it is best to condition decisions on as much information as is available (D vs. B3).
\end{itemize}

The absolute results are also quite promising, in that this type of
parsing has state-of-the-art accuracy.  This may be accurate enough
for the parser to be useful as part of a larger system.  It is
striking that these results are obtained with such simple models.  For
example, there is no special treatment of NP chunks, verbs, or
punctuation.

Third, we have tried to provide enough details so that others can
either replicate our experimental work or improve on our design
without having to reinvent it first.  The present work is a modest
entry in the relatively new area of experimentally comparing
statistical NLP methods (e.g.,
\cite{caraballo+charniak,chen+goodman,mooney}).  Two experimental
practices described herein have not, to our knowledge, been previously
applied in the comparison of parsers: an evaluation metric based on
dependency attachments (as proposed by \cite{lin95}), and the use of
non-parametric methods to evaluate statistical significance.

\end{document}